\documentclass{llncs}

\usepackage{tikz} 
\usepackage{multirow} 
\usepackage{graphicx} 

\usepackage{booktabs}
\usepackage{siunitx}

\usepackage[colorinlistoftodos]{todonotes}

\usepackage{subfig}

\begin{document}

\title{Evaluating surgical skills from kinematic data using convolutional neural networks}
\titlerunning{CNN on JIGSAWS}  
%
\author{Hassan Ismail Fawaz, 
Germain Forestier, 
Jonathan Weber, \\
Lhassane Idoumghar \and
Pierre-Alain Muller
}

\authorrunning{Ismail Fawaz et al.} 

%
\institute{
IRIMAS, Universit\'{e} de Haute-Alsace,
68100 Mulhouse, France\\
\email{\{first-name\}.\{last-name\}@uha.fr}
}

\maketitle              

\begin{abstract}
The need for automatic surgical skills assessment is increasing, especially because manual feedback from senior surgeons observing junior surgeons is prone to subjectivity and time consuming. 
Thus, automating surgical skills evaluation is a very important step towards improving surgical practice. 
In this paper, we designed a Convolutional Neural Network (CNN) to evaluate surgeon skills by extracting patterns in the surgeon motions performed in robotic surgery. 
The proposed method is validated on the JIGSAWS dataset and achieved very competitive results with 100\% accuracy on the suturing and needle passing tasks. 
While we leveraged from the CNN’s efficiency, we also managed to mitigate its black-box effect using class activation map. 
This feature allows our method to automatically highlight which parts of the surgical task influenced the skill prediction and can be used to explain the classification and to provide personalized feedback to the trainee.

\keywords{kinematic data, RMIS, deep learning, CNN}
\end{abstract}

\section{Introduction}

Over the past one hundred years, the classical training program of Dr. William Halsted has governed surgical training in different parts of the world~\cite{polavarapu2013100}.
His teaching philosophy of ``see one, do one, teach one'' is still one of the most practiced methods to this day~\cite{ahmidi2017a}.
The idea is that the trainee could become an expert surgeon by watching and assisting in mentored surgeries~\cite{polavarapu2013100}.
These training methods, although broadly used, lack of an objective surgical skill evaluation technique~\cite{kassahun2016surgical}. 
Conventional surgical skill assessment is currently based on checklists that are filled by an expert surgeon observing the surgery~\cite{tedesco2008simulation}.
In an attempt to evaluate surgical skills without relying on an expert's opinion, Objective Structured Assessment of Technical Skills (OSATS) has been proposed and is used in clinical practice~\cite{niitsu2013using}.
Unfortunately, this type of observational evaluation is still prone to several external and subjective variables: the checklists' development process, the inter-rater reliability and the evaluator bias~\cite{hatala2015constructing}. 

Other studies showed that a strong relationship exists between the postoperative outcomes and the technical skill of a surgeon~\cite{bridgewater2003surgeon}. 
This type of approach suffers from the fact that a surgery's outcome also depends on the patient's physiological characteristics~\cite{kassahun2016surgical}. 
In addition, acquiring such type of data is very difficult, which makes these skill evaluation methods difficult to apply for surgical training.
Recent advances in surgical robotics such as the \emph{da Vinci} surgical robot (Intuitive Surgical Inc. Sunnyvale, CA) enabled the collection of motion and video data from different surgical activities.
Hence, an alternative for checklists and outcome-based methods is to extract, from these motion data, global movement features such as the surgical task's time completion, speed, curvature, motion smoothness and other holistic features~\cite{fard2018automated,zia2017automated,kassahun2016surgical}. 
Although most of these methods are efficient, it is not clear how they could be used to provide a detailed and constructive feedback for the trainee to go beyond the simple classification into a category (i.e. novice, expert, etc.). 
This is problematic as studies \cite{islam2016affordable} showed that feedback on medical practice allows surgeons to improve their performance and reach higher skill levels.

Recently, a new field named \emph{Surgical Data Science}~\cite{maier-hein2017surgical} has emerged thanks to the increasing access to large amounts of complex data which pertain to the patient, the staff and sensors for perceiving the patient and procedure related data such as videos and kinematic variables~\cite{gao2014jhu}.
As an alternative to extracting global movement features, recent studies tend to break down surgical tasks into smaller segments called surgical gestures, manually before the training phase, and assess the performance of the surgical task based on the assessment of these gestures~\cite{lingling2012sparse}. 
Although these methods obtained very accurate and promising results in terms of surgical skill evaluation, they require a huge amount of labeled gestures for the training phase~\cite{lingling2012sparse}.
We have identified two main limits in the existing approaches that classify a surgeon's skill level based on the kinematic data.
First is the lack of an interpretable result of skill evaluation usable by the trainee to achieve higher skill levels.
Additionally current state of the art Hidden Markov Models require gesture boundaries that are pre-defined by human annotators which is time consuming and prone to inter-annotator reliability~\cite{vedula2016analysis}. 

\input{input/fig-tikz}
In this paper, we propose a new architecture of Convolutional Neural Networks (CNN) dedicated to surgical skill evaluation (Figure~\ref{fig:archi}).
By using one dimensional filters over the kinematic data, we mitigate the need to pre-define sensitive and unreliable gesture boundaries.
The original hierarchical structure of our deep learning model enables us to represent the gestures in latent low-level variables (first and second layers), as well as capturing global information related to the surgical skill level (third layer).
To provide interpretable feedback, instead of using a final fully-connected layer like most traditional approaches~\cite{zhou2016learning}, we place a Global Average Pooling (GAP) layer which enables us to benefit from the Class Activation Map~\cite{zhou2016learning} (CAM) to visualize which parts of the surgical task contributed the most to the surgical skill classification (Figure~\ref{fig:trials}).
We demonstrate the accuracy of our approach using a standardized experimental setup on the largest publicly available dataset for surgical data analysis: the JHU-ISI Gesture and Skill Assessment Working Set (JIGSAWS)~\cite{gao2014jhu}.
The main contribution of our work is to show that deep learning can be used to understand the latent and complex structures of what constitutes a surgical skill, especially that there is still much to be learned on what is exactly a surgical skill~\cite{kassahun2016surgical}.

\section{Method}

\subsection{Dataset}
We first present briefly the dataset used in this paper as we rely on features definition to describe our method.
The JIGSAWS~\cite{gao2014jhu} dataset has been collected from eight right-handed subjects with three different skill levels (Novice (N), Intermediate (I) and Expert (E)) performing three different surgical tasks (suturing, needle passing and knot tying) using the \emph{da Vinci} surgical system.
Each subject performed five trials of each task.
For each trial the kinematic and video data were recorded.

In our work, we only focused on kinematic data which are numeric variables of four manipulators: left and right masters (controlled directly by the subject's hands) and left and right slaves (controlled indirectly by the subject via the master manipulators). 
These kinematic variables (76 in total) are captured at a frequency equal to 30 frames per second for each trial.  
We considered each trial as a multivariate time series (MTS) and designed a one dimensional CNN dedicated to learn automatically useful features for surgical skill classification.   

\subsection{Architecture}
Our approach takes inspiration of the recent success of CNN for time series classification~\cite{wang2017time}. 
The proposed architecture (Figure~\ref{fig:archi}) has been specifically designed to classify surgical skills using kinematic data. 
The input of the CNN is a MTS with variable length $l$ and 76 channels.
The output layer contains the surgical skill level (N, I, E).
Comparing to CNNs for image classification, where usually the network's input has two dimensions (width and height) and 3 channels (RGB), our network's input is a time series with one dimension (length $l$ of the surgical task) and 76 channels (the kinematic variables $x,y,z,x^\prime$, etc.).

The main challenge we encountered when designing our network was the huge number of input channels (76) compared to the RGB channels (3) for the image classification task. 
Therefore, instead of applying the convolutions over the 76 channels, we proposed to carry out different convolutions for each cluster and sub-cluster of channels.
In order to decide which channels should be grouped together, we used domain knowledge when clustering the channels.

First we divide the 76 variables into four different clusters, such as each cluster contains the variables from one of the four manipulators: the $1^{st},2^{nd},3^{rd}$ and $4^{th}$ clusters correspond respectively to the four manipulators (ML: master left, MR: master right, SL: slave left and SR: slave right) of the \emph{da Vinci} surgical system. 
Thus, each cluster contains 19 of the 76 total kinematic variables.  

Next, each cluster of 19 variables is split into five different sub-clusters such as each sub-cluster contains variables that we hypothesize are highly correlated.
For each cluster, the variables are grouped into five sub-clusters: 
$1^{st}$ sub-cluster with 3 variables for the Cartesian coordinates ($x,y,z$);
$2^{nd}$ sub-cluster with 3 variables for the linear velocity ($x^{\prime},y^{\prime},z^{\prime}$);
$3^{rd}$ sub-cluster with 3 variables for the rotational velocity ($\alpha^{\prime},\beta^{\prime},\gamma^{\prime}$);
$4^{th}$ sub-cluster with 9 variables for the rotation matrix R;
$5^{th}$ sub-cluster with 1 variable for the gripper angular velocity ($\theta$).  

Figure~\ref{fig:archi} shows how the convolutions in the first layer are different for each sub-cluster of channels. 
Following the same reasoning, the convolutions in the second layer are different for each cluster of channels (ML, MR, SL and SR). 
However, in the third layer, the same convolutions are applied for all channels. 

In order to reduce the number of parameters in our model and benefit from the CAM method~\cite{zhou2016learning}, we replaced the fully-connected layer with a GAP operation after the third convolutional layer. 
This results in a summarized MTS that shrinks from a length $l$ to 1, while preserving the same number of channels in the third layer.
As for the output layer, we use a fully-connected softmax layer with three neurons, one for each class (N, I, E).

Without any cross-validation, we choose to use $8$ filters at the first convolutional layer, then we increase the number of filters (by a factor of $2$), thus balancing the number of parameters for each layer while going deeper into the network. 
The Rectified Linear Unit (ReLU) activation function is employed for the three convolutional layers with a filter size of $3$ and a stride of $1$. 

\subsection{Training \& Testing}
To train the network, we used the multinomial cross-entropy as our objective cost function. 
The network's parameters were optimized using Adam~\cite{kingma2015adam}. 
Following~\cite{wang2017time}, without any fine-tuning, the learning rate was set to $0.001$ and the exponential decay rates of the first and second moment estimates were set to $0.9$ and $0.999$ respectively.
Each trial was used in a forward-pass followed by a back-propagation update of the weights which were initialized using Glorot's uniform initialization~\cite{glorot2010understanding}.
Before each training epoch, the train set was randomly shuffled.
We trained the network for $1000$ epochs, then by saving the model at each training epoch, we chose the one that minimized the objective cost function on a random (non-seen) split from the training set.
Thus, we only validate the number of epochs since no extra-computation is needed to perform this step. 
Finally, to avoid overfitting, we added a $l2$ regularization parameter equal to $10^{-5}$. 
  Since we did not fine-tune the model's hyper-parameters, the same network architecture with the same hyper-parameters was trained on each surgical task resulting in three different models\footnote{\scriptsize Our source code is available on \url{https://germain-forestier.info/src/miccai2018/}}.

To evaluate our approach we adopted the standard benchmark configuration, Leave One Super Trial Out (LOSO)~\cite{ahmidi2017a}: for each iteration of cross-validation (five in total), one trial of each subject was left out for the test and the remaining trials were used for training.

\subsection{Class Activation Map}\label{sec:cam}
By employing a GAP layer, we benefit from the CAM~\cite{zhou2016learning} method, which makes it possible to identify which regions of the surgical task contributed the most to a certain class identification. 
Let $A_k(t)$ be the result of the third convolutional layer which is a MTS with $K$ channels (in our case $K$ is equal to 32 filters and $t$ denotes the time dimension). 
Let $w_k^c$ be the weight between the output neuron of class $c$ and the $k^{th}$ filter. 
Since a GAP layer is used, the input to the output neuron of class $c$ ($z_c$) and the CAM ($M_c(t)$) can be defined as:
\begin{equation}\label{eq-1}
z_c=\sum_k{w_k^c\sum_t{A_k(t)}}=\sum_t{\sum_k{w_k^c A_k(t)}} \quad ; \quad  M_c(t)=\sum_k{w_k^c A_k(t)}
\end{equation}
In order to avoid upsampling the CAM, we padded the input of each convolution with zeros, thus preserving the initial MTS length $l$ throughout the convolutions. 

\section{Results}
\subsection{Surgical skill classification}
Table~\ref{table:results} reports the micro and macro measures (defined in~\cite{ahmidi2017a}) of four different methods for the surgeons' skill classification of the three surgical tasks. 
For our approach (CNN), we report the average of 40 runs to eliminate any bias due to the random seed. 
From these results, it appears that the CNN method is much more accurate than the other approaches with 100\% accuracy for the suturing and needle passing tasks.
As for the knot tying task, we report 92.1\% and 93.2\% respectively for the micro and macro configurations. 
Indeed, for knot tying, the model is less accurate compared to the other two.
This is due to the complexity of this task, which is in compliance with the results of the other approaches.  

In~\cite{lingling2012sparse}, the authors designed Sparse Hidden Markov Models (S-HMM) to evaluate the surgical skills. 
Although the latter method utilizes the gesture boundaries during the training phase, our approach achieves much higher accuracy while still providing the trainee with interpretable skill evaluation. 

Approximate Entropy (ApEn) is used to extract features from each trial~\cite{zia2017automated}, which are then fed to a nearest neighbor classifier. 
Although both methods (ApEn and CNN) achieve state of the art results with 100\% accuracy for the suturing and needle passing surgical tasks, it is not clear how ApEn could be extended to provide feedback for the trainee. 
In addition, we hypothesize that by doing cross-validation and hyper-parameters fine tuning, we could squeeze higher accuracy from the CNN, especially for the knot tying task.

Finally, in~\cite{forestier2017discovering}, the authors introduce a sliding window technique with a discretization method to transform the MTS into bag of words. 
Then, they build a vector for each class from the frequency of the words, which is compared to vectors of the MTS in the test set to identify the nearest neighbor with a cosine similarity metric.   
The authors emphasized the need to obtain \emph{interpretable} surgical skill evaluation, which justified their relatively low accuracy. 
On contrast, our approach does not trade off accuracy for feedback: CNN is much more \emph{accurate} and equally \emph{interpretable}.   
\begin{table}
	\centering
	\caption{Surgical skill classification results (\%)
    }
	\begin{tabular}{l|cc|cc|cc}
		\toprule
		\multirow{2}{*}{\scriptsize Method} &
		\multicolumn{2}{c}{\scriptsize Suturing} &
		\multicolumn{2}{c}{\scriptsize Needle Passing} &
		\multicolumn{2}{c}{\scriptsize Knot Tying} \\
		& {\scriptsize Micro} & {\scriptsize Macro} & {\scriptsize Micro} & {\scriptsize Macro} & {\scriptsize Micro} & {\scriptsize Macro} \\
		\midrule
		S-HMM~\cite{lingling2012sparse} & 97.4 & n/a  & 96.2 & n/a & 94.4 & n/a \\
		ApEn~\cite{zia2017automated} & \textbf{100} & n/a & \textbf{100} & n/a & \textbf{99.9} & n/a \\
		Sax-Vsm~\cite{forestier2017discovering} & 89.7 & 86.7 & 96.3 & 95.8 & 61.1 & 53.3 \\
		CNN (proposed) & \textbf{100} & \textbf{100} & \textbf{100} & \textbf{100} & 92.1 & \textbf{93.2} \\
		\bottomrule
	\end{tabular}
    \label{table:results}
\end{table}

\begin{figure}
\centering
    \subfloat[The last frame of subject (Novice) H's fourth trial of the suturing task.]{
    
 \includegraphics[width=.35\linewidth]{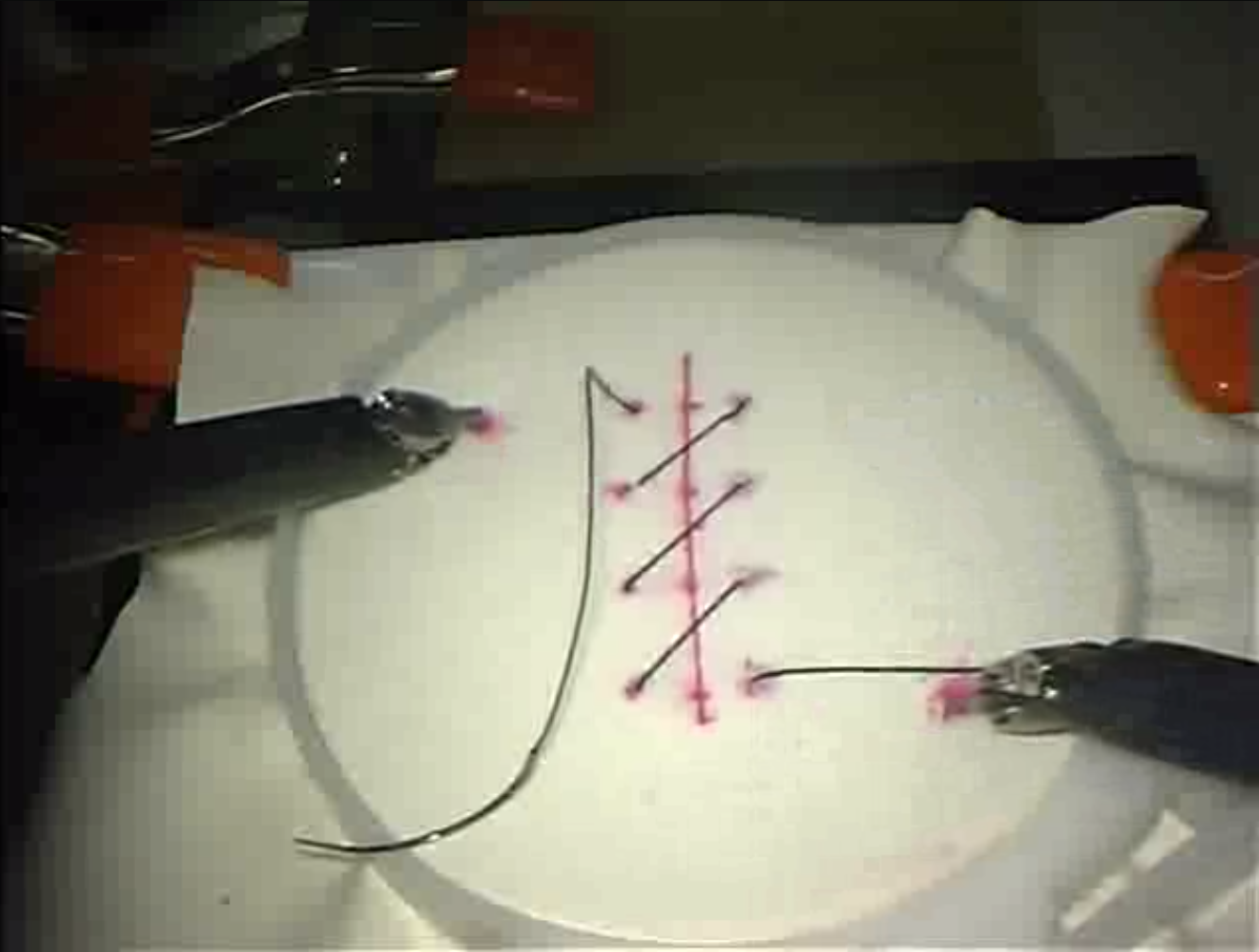}
      \label{sub:screenshot}}
 \hspace{.1cm}
    \subfloat[Trial's corresponding trajectory for the left master manipulator (best viewed in color).]{
 \includegraphics[height=0.235\textheight,width=.35\linewidth]{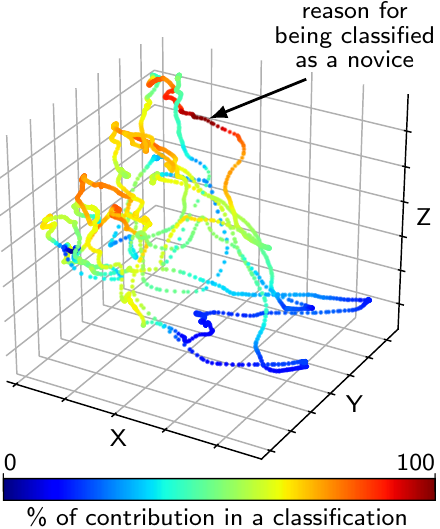}
      \label{sub:feedback}
      }
    \caption{Example of feedback using Class Activation Map (a video illustrating this feedback is available on \url{https://germain-forestier.info/src/miccai2018/}).
    }
    \label{fig:trials}
\end{figure}

\subsection{Feedback visualization}\label{sec:feedback}
The CAM technique allows us to visualize which parts of the trial contributes the most to a certain skill classification. 
Patterns in movements could be understood by identifying for example discriminative behaviors specific to a novice or an expert.
We can also pinpoint to the trainees their good/bad movements in order to improve themselves and achieve potentially higher skill levels.  

Figure~\ref{fig:trials} gives an example on how to visualize the feedback for the trainee by constructing a heatmap from the CAM. 
A trial of a novice subject is studied: its last frame is shown in Figure~\ref{sub:screenshot} and its corresponding heatmap is illustrated in Figure~\ref{sub:feedback}. 
In the latter, the model was able to detect which movements (red area) were the main reason behind subject H's classification (as a novice). 
This feedback could be used to explain to a young surgeon which movements are classifying him/her as a novice and which ones are classifying another subject as an expert. 
Thus, the feedback could guide the novices into becoming experts.

\section{Conclusion}
In this paper, we presented a new method for classifying surgical skills. 
By designing a specific CNN, we achieved 100\% accuracy, while providing interpretability that justifies a certain skill evaluation, which reduces the CNN's black-box effect.

In our future work, due to the natural extension of CNNs to image classification, we aim at developing a unified CNN framework that uses both video and kinematic data to classify surgical skills accurately and to provide highly interpretable feedback for the trainee.

%
%
\bibliographystyle{splncs03}
\bibliography{biblio}

\end{document}